\documentclass[runningheads]{llncs}

\usepackage[T1]{fontenc}

\usepackage{graphicx}

\usepackage{authblk}
\usepackage{pgf-pie}
\usepackage{caption}
\usepackage{algpseudocode}
\usepackage{xcolor}
\usepackage{adjustbox}
\usepackage{subfig}
\usepackage{hyperref}

\definecolor{sb1}{HTML}{66C2A5}
\definecolor{sb2}{HTML}{FC8D62}
\definecolor{sb3}{HTML}{8DA0CB}
\definecolor{sb4}{HTML}{E78AC3}
\definecolor{sb5}{HTML}{FFD92F}

\title{Data Efficient Training of a U-Net Based Architecture for Structured Documents Localization}

\titlerunning{SDL-Net}

\author{
    Anastasiia Kabeshova\inst{1} \and Guillaume Betmont\inst{1} \and Julien Lerouge\inst{1}  \and \newline Evgeny Stepankevich\inst{1} \and Alexis Bergès\inst{1}
}

\authorrunning{A. Kabeshova et al.}

\institute{
    QuickSign, Paris, France\\
    \email{\{anastasiia.kabeshova,guillaume.betmont,julien.lerouge,\\evgeny.stepankevich,alexis.berges\}@quicksign.com}\\
    \url{https://www.quicksign.com/}
}

\usepackage{algorithm}
\usepackage{algpseudocode}

\begin{document}
\setcounter{tocdepth}{2}
\maketitle

\begin{abstract}
Structured documents analysis and recognition are essential for modern online on-boarding processes, and document localization is a crucial step to achieve reliable key information extraction. While deep-learning has become the standard technique used to solve document analysis problems, real-world applications in industry still face the limited availability of labelled data and of computational resources when training or fine-tuning deep-learning models. To tackle these challenges, we propose SDL-Net: a novel U-Net like encoder-decoder architecture for the localization of structured documents. Our approach allows pre-training the encoder of SDL-Net on a generic dataset containing samples of various document classes, and enables fast and data-efficient fine-tuning of decoders to support the localization of new document classes. We conduct extensive experiments on a proprietary dataset of structured document images to demonstrate the effectiveness and the generalization capabilities of the proposed approach.
\end{abstract}

\section{Introduction}\label{sec:Introduction}

\begin{figure}[ht]
    \centering
     \subfloat[Input]{\includegraphics[width=.25\linewidth]{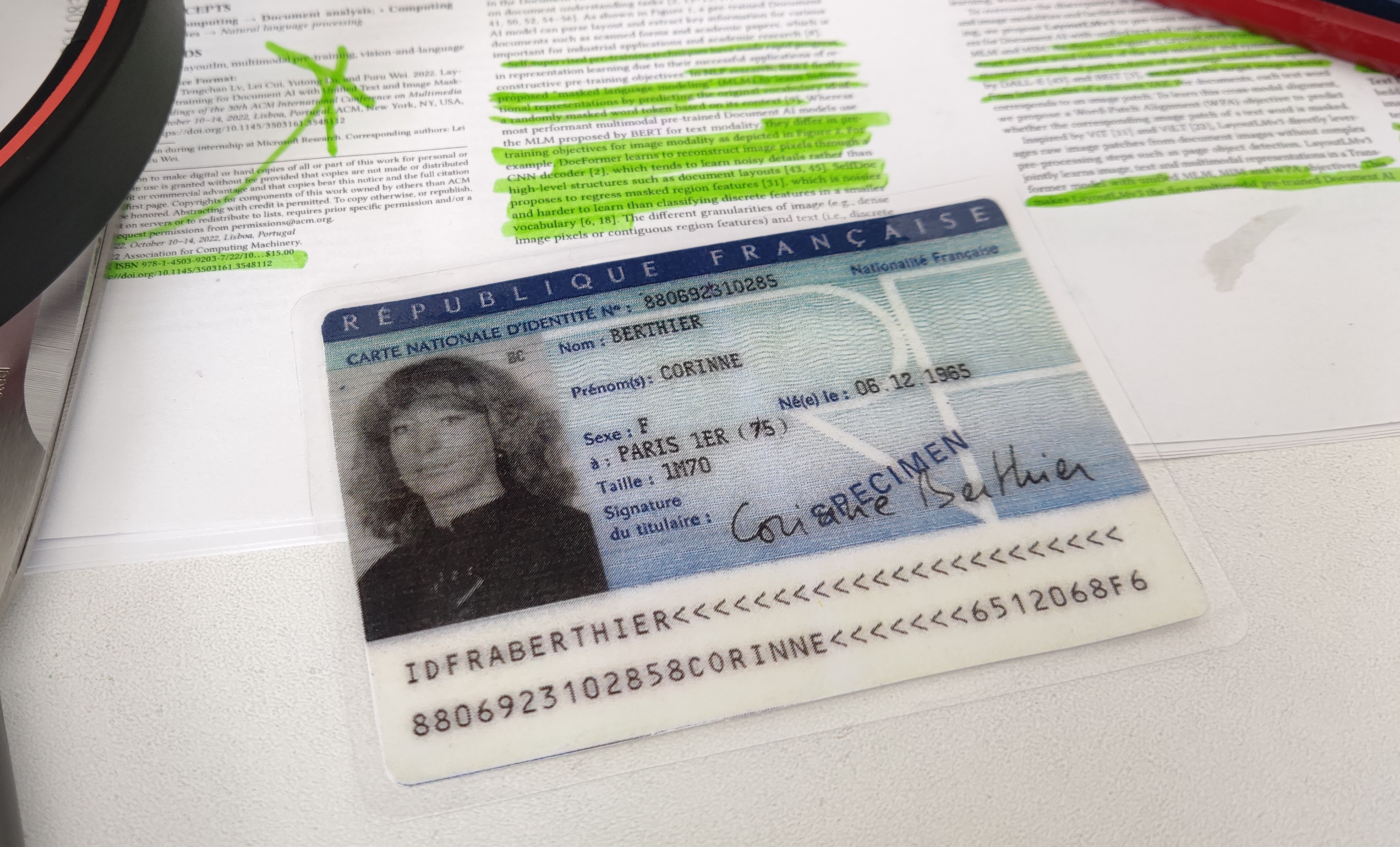}\label{fig:document_localization_a}} \quad
     \subfloat[Heat maps]{\includegraphics[width=.15\linewidth]{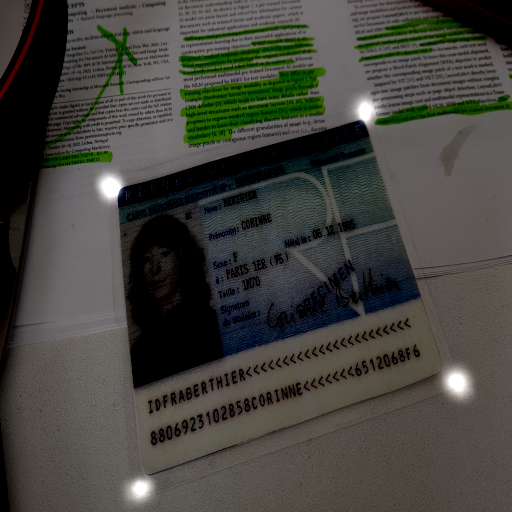}\label{fig:document_localization_b}} \quad
     \subfloat[Quadrangle]{\includegraphics[width=.25\linewidth]{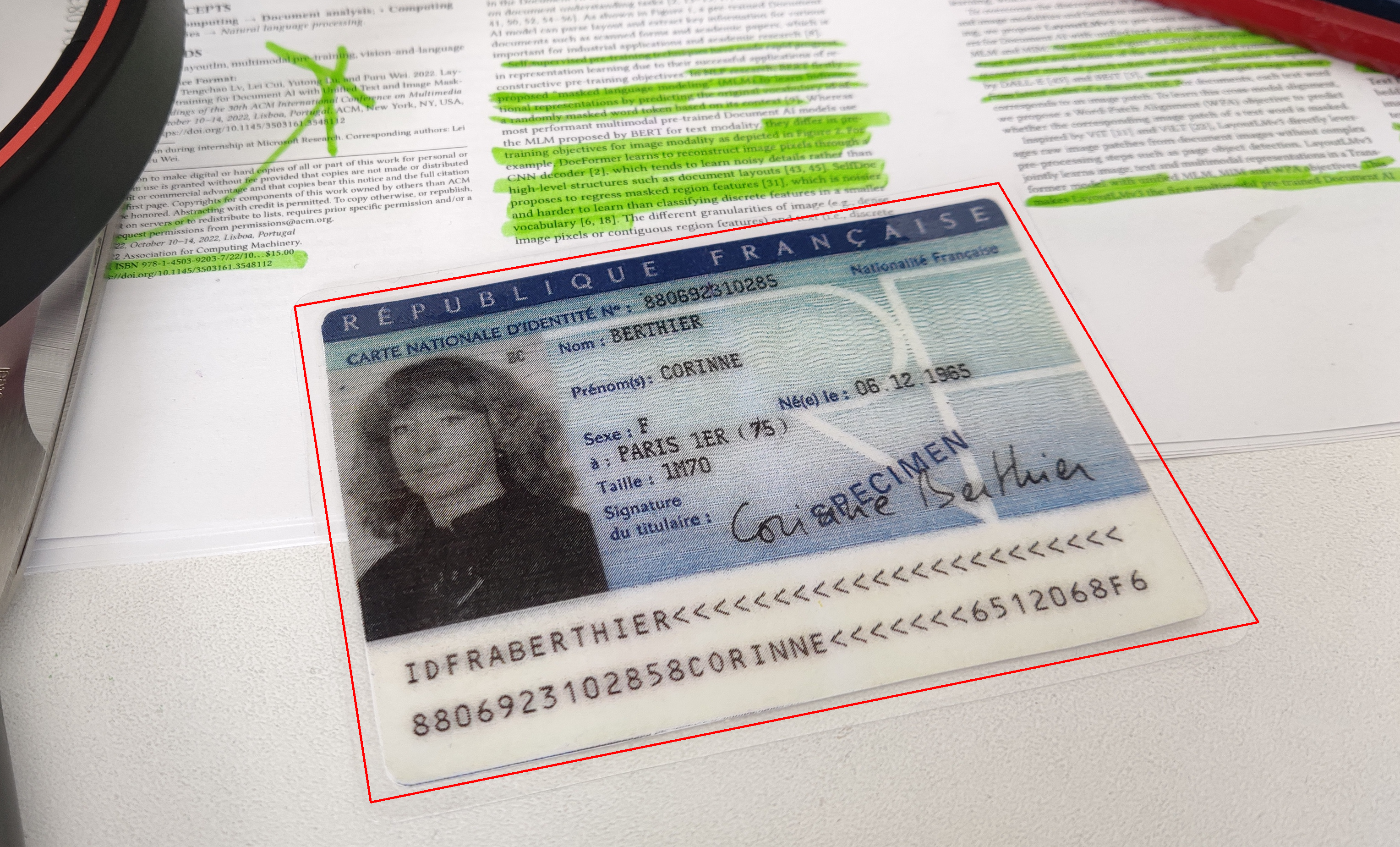}\label{fig:document_localization_c}} \quad
     \subfloat[Rectification]{\includegraphics[width=.22\linewidth]{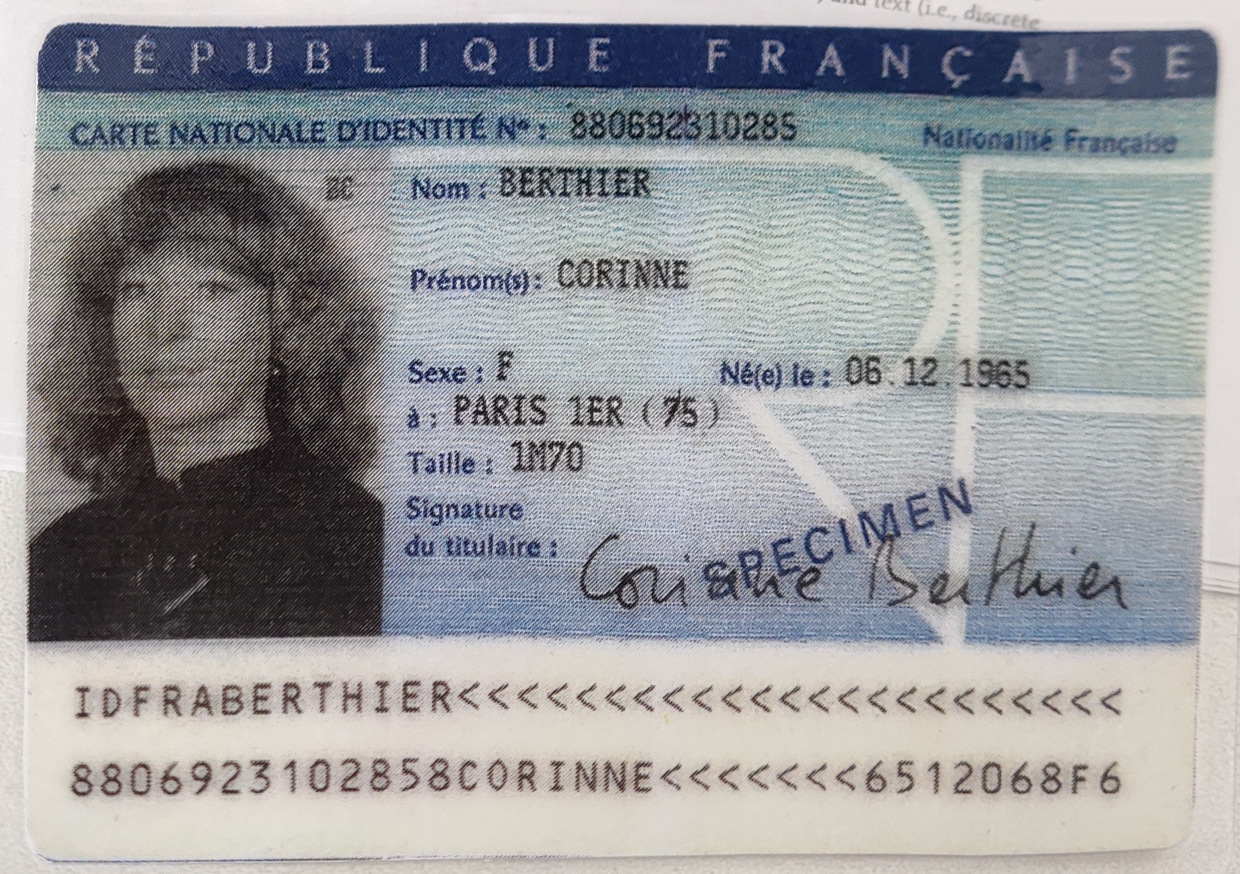}\label{fig:document_localization_d}} \\
    \ \\
    \caption{Proposed document localization approach. The input image (a) is resized to 512x512 pixels, and processed by a U-Net with 4 output channels, which tries to predict the four corners location. From the output heatmaps (b) (they are merged and blended on the resized image for better visualization), the corners are detected and a quadrangle (c) is constructed to represent the position of the document. Finally, a homography between the quadrangle (c) and a rectangle obtained by estimating the document ratio aspect is computed to obtain the rectified image (d).}
    \label{fig:document_localization}
\end{figure}

Many on-boarding processes require the verification of information about an applicant, often for legal or regulatory reasons, by automatically processing images of structured documents (such as identity cards, passports or driving licenses). The significant growth of online operations in the past decade requires the processing of document images to be fast and accurate. Also, the preferred acquisition modality has quickly transitioned from personal scanners to smartphone cameras. Hence, extracting documents from backgrounds and correcting the perspective have become important challenges\cite{burie2015icdar2015}. It's possible to obtain a straightened image, as shown in Figure~\ref{fig:document_localization_d}, of a document by predicting its precise localization. This straightening operation helps text and graphical objects detection as well as text recognition.

In this work, we explore whether an encoder-decoder architecture can help us tackle an industrial constraint, namely the issue of having large enough labeled datasets to obtain reasonable performances, for the document localization task, on classes of documents for which acquiring many sample documents is difficult, e.g. for privacy, legal, or geographical reasons, and  labeling many documents is expensive.



For this, we introduce a model architecture inspired by U-Net \cite{ronneberger2015u} for structured documents localization, that takes advantage of the encoder-decoder approach in order to learn strong intermediary representations in the encoder part, and to reduce the need for large datasets during finetuning of the decoder part. By building a generic encoder, we manage to fine-tune specific decoders for each document class to be supported with small amounts of labelled images, and consequently a relatively small computational footprint and a low labeling cost. We conduct extensive experiments on an internal proprietary dataset of structured document images to demonstrate the effectiveness and the generalization capabilities of the proposed approach.

\section{Related Work}\label{sec:Review}

\subsection{Document Localization}

Research has been conducted on document localization since the start of camera-based document processing development. During the 2000s were developed the first methods \cite{clark2002recovery,lampert2005oblivious,stamatopoulos2007automatic} to track rectangular frames in camera image, appearing as quadrilaterals, in order to correct the perspective before an OCR processing. In the recent literature about document localization, four main types of approaches to solve document localization are described : contour detection methods, keypoints detection methods, template matching methods and segmentation methods.

\textbf{Contour detection methods}, such as \cite{zhukovsky2017segments,puybareau2018real,tropin2021approach,tropin2021improved}, follow the idea of traditional image processing methods, which used to detect the edges of documents using edge detectors (like Canny edge detector, or Holistically-Nested Edge Detector \cite{xie2015holistically}), mathematical morphology, and line or segment detectors (like Hough transform, or Line Segment Detector \cite{von2008lsd}). These methods are robust to partial document occlusion (missing corner), but are rather sensitive to bad lighting conditions and bad contrast. Also, they can't, in general, directly predict the orientation of the document.

\textbf{Corner detection methods}, such as \cite{javed2017real,zhu2019coarse}, detect the four corners of rectangular documents to estimate the perspective and rectify the image. These methods are not robust to partial occlusion, but can directly predict the orientation of the document when the corners are labeled. Our approach belongs to this category.

\textbf{Template matching methods}, such as \cite{awal2017complex,chiron2021id,matalov2021rfdoc}, need a correct classification of the document image, which may be obtained before the template matching step, or simulateously. A matching criterion between the query image and the template images is needed to estimate a transformation between the query space and the template space. These criterion are often based on keypoints detectors and descriptors. Document orientation is given by the matching result.

\textbf{Segmentation methods}, such as \cite{xu2016hierarchical,leal2016smartphone,castelblanco2020machine,das2020hu}, treat document localization as a segmentation task. A pixel-level mask of the document is regressed, and then transformed to a quadrilateral, to estimate the position. This also need further post-processing, like contour-based approaches, to predict orientation of the document.

Some hybrid approaches have also been tested, such as \cite{skoryukina2018document} which combines template-matching techniques with edge detection, or, more recently, \cite{wu2023ldrnet} which combines corners prediction, edges prediction and classification in a single lightweight CNN model. Some other approaches, such as \cite{sheshkus2019houghnet}, solve the perspective correction issue directly, without estimating the document position.

\subsection{Encoder-Decoder Architectures}

An encoder is a subset of a deep neural network, whose task is to produce a high-level features of the input images, generally in feature maps with reduced spatial dimension but with a lot more channels. A decoder takes the feature maps of the encoder as input, and transforms it into the required target output. In image processing field, popular convolutional neural network (CNN) such as VGG \cite{simonyan2014very} or ResNet \cite{he2016deep} are commonly used as backbones of the encoder, and their weights are typically pre-trained on large datasets, like ImageNet \cite{russakovsky2015imagenet}. Some relatively recent deep architectures, like Fully Convolutional Networks (FCN) \cite{long2015fully}, U-Net \cite{ronneberger2015u} and SegNet \cite{badrinarayanan2017segnet}, have dramatically improved performance on pixel-wise image-to-image learning tasks, including semantic segmentation or edge detection \cite{ji2021cnn}, by learning to decode low-resolution feature maps to pixel-wise predictions. In other fields, encoder-decoder architectures have also been applied successfully, like BERT \cite{devlin2018bert} for Natural Language Processing (NLP).
\section{Methodology}\label{sec:methodology}

\subsection{Model architecture}

\begin{figure}[ht]
  \centering
  \includegraphics[width=.6\linewidth]{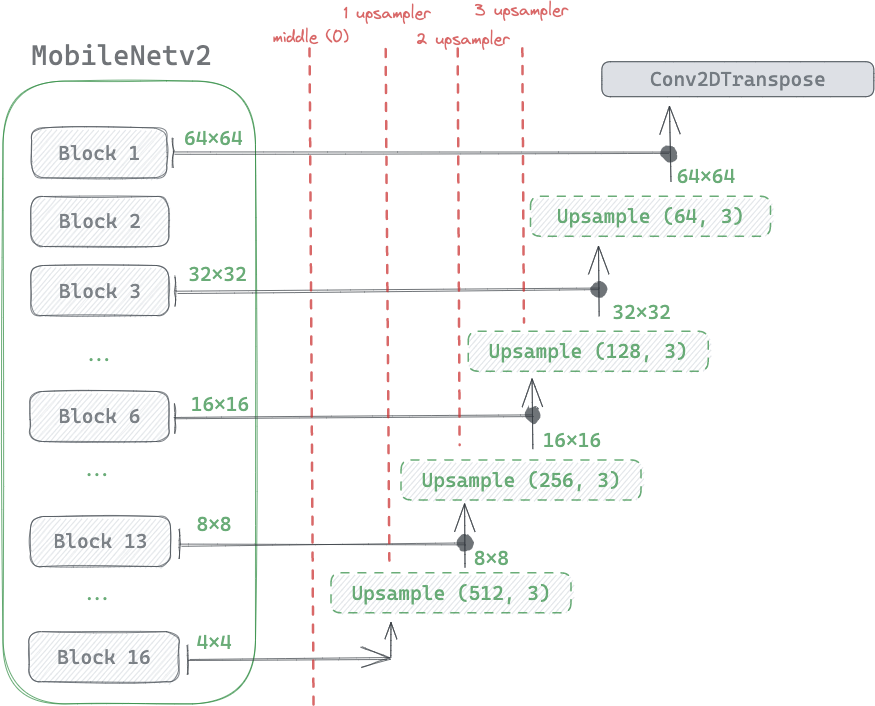}
  \caption{U-Net architecture, using MobileNetV2 as backbone. Dotted red lines represent split candidates.}
  \label{fig:unet_with_split}
\end{figure}

We implement a model based on the U-Net \cite{ronneberger2015u} architecture for key-point localization, as presented in Figure~\ref{fig:unet_with_split}. We considered four possible splits between encoder and decoder parts, in order to evaluate the trade-off between freezing more weights in the encoder part opposed to having more free weights in the decoder part during fine-tuning. We limit ourselves to rectangular documents, so we need to predict the position of four corners only. This covers all structured document classes that we are confronted with in practice. In the rest of our article, the number of output channels is fixed to four (one per corner).

Similarly to LDRNet \cite{wu2023ldrnet}, we used a MobileNetV2 \cite{sandler2018mobilenetv2} as the backbone of our model, in order to reduce the number of trainable parameters and memory footprint in comparison to the original U-Net model. This backbone is pre-trained on ImageNet. We added skip connections between the backbone blocks and the upsamplers, as in U-Net, to help reconstructing the full resolution outputs.


\subsection{Dataset}

\begin{figure}[ht]
  \centering
  \subfloat[ID]{\includegraphics[width=0.15\textwidth]{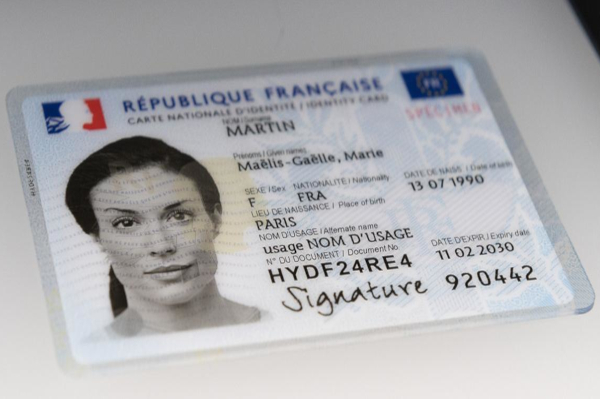}\label{fig:document_examples_a}}
  \quad
  \subfloat[DL]{\includegraphics[width=0.15\textwidth]{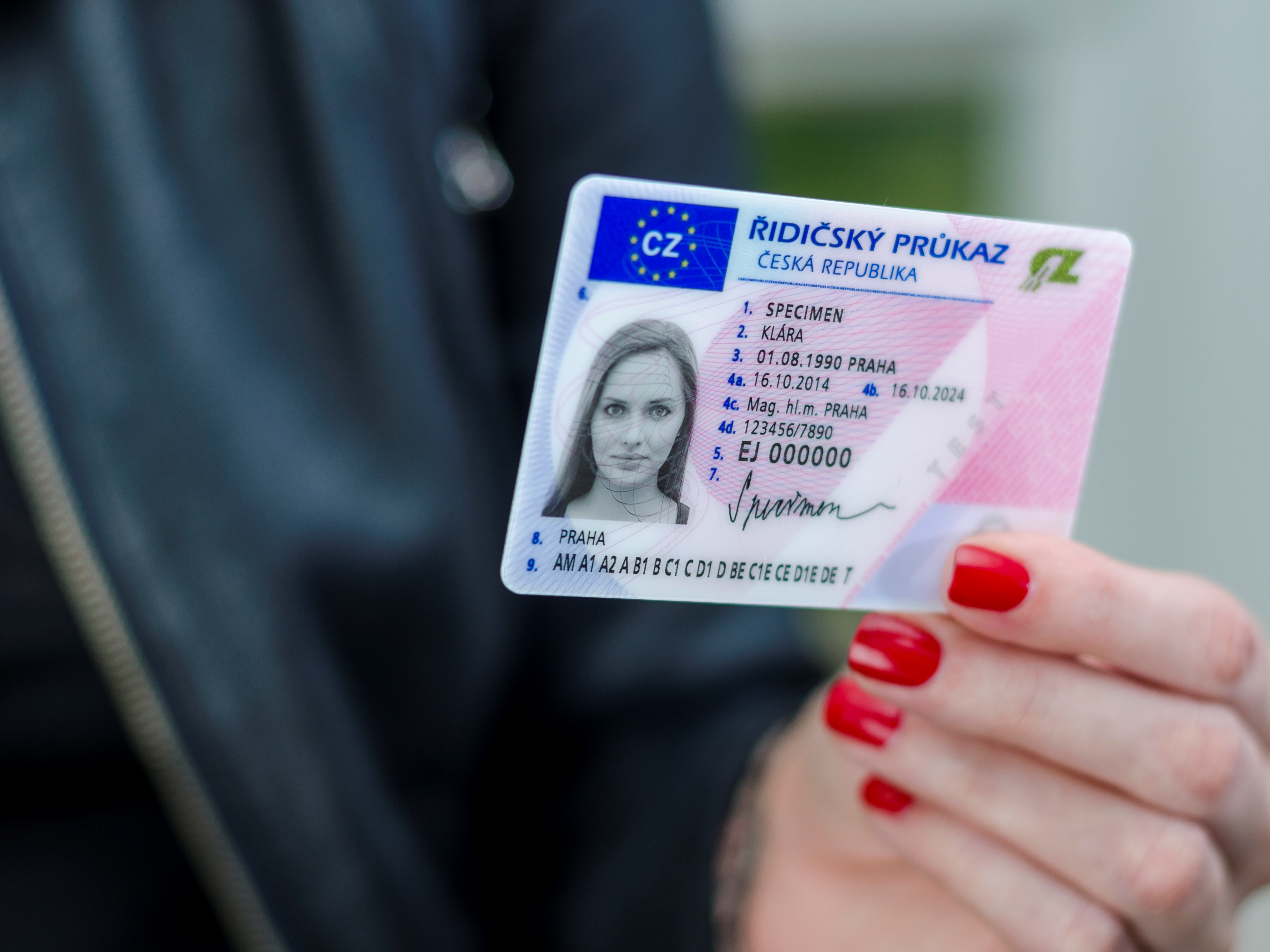}\label{fig:document_examples_b}}
  \quad
  \subfloat[P]{\includegraphics[width=0.15\textwidth]{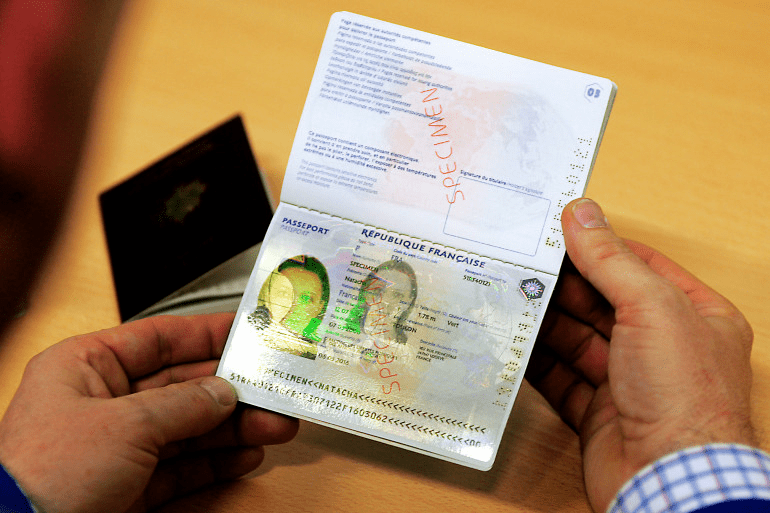}\label{fig:document_examples_c}}
  \quad
  \subfloat[RP]{\includegraphics[width=0.15\textwidth]{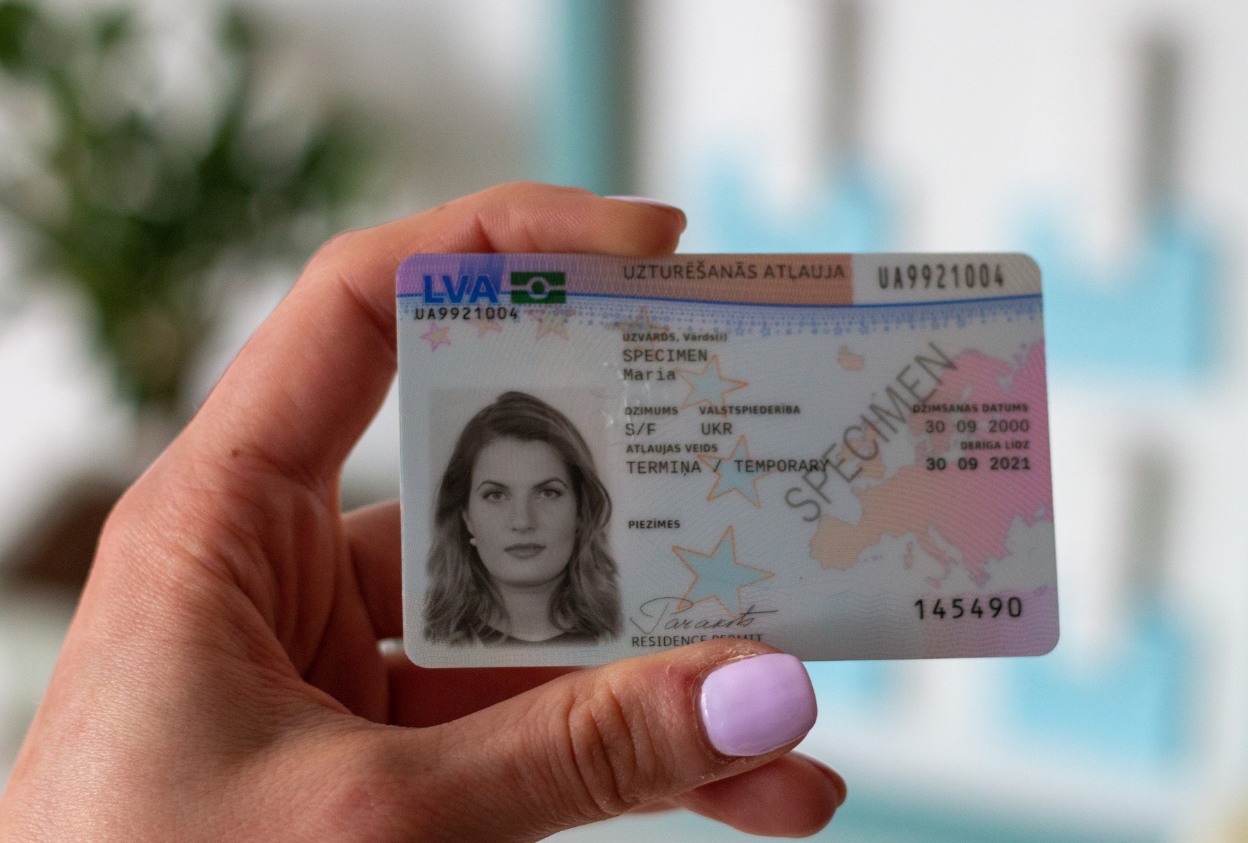}\label{fig:document_examples_d}}
  \quad
  \subfloat[VRC]{\includegraphics[width=0.15\textwidth]{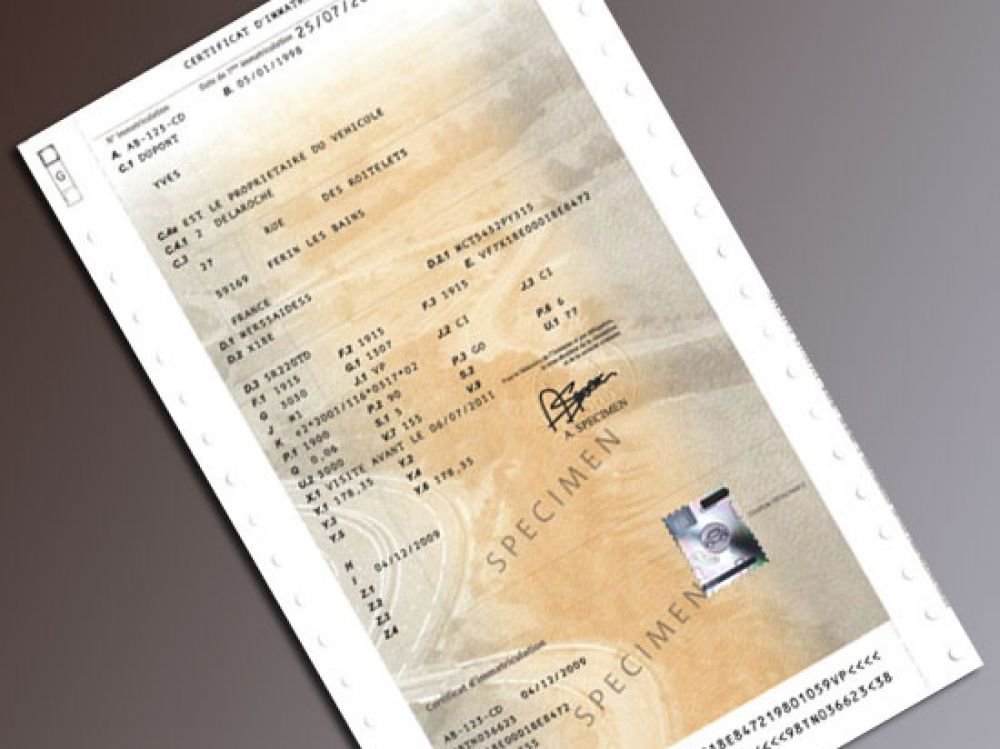}\label{fig:document_examples_e}}
  \quad \\
  \quad \\
  \caption{Specimens of the five document classes.}
  \label{fig:document_examples}
\end{figure}

We randomly selected 8,942 images from a private dataset of document images. This dataset contains personal information and cannot be published, to our regret. Each of these images belongs to one of the following five classes: Identity Card (ID, 21\% of the dataset), Driving Licence (DL, 19\%), Passport (P, 25\%), Residence permit (RP, 14\%), or Vehicle Registration Certificate (VRC, 21\%). Examples of each document class are presented in Figure~\ref{fig:document_examples}. We consider that each of those document classes constitutes a separate dataset, and we randomly split each of those five datasets into train (70\%), validation (15\%) and test (15\%) sub-sets. All images were labelled manually, using LabelStudio \cite{tkachenko2020label}, by determining the position of the four corners, namely top-left (TL), top-right (TR), bottom-left (BL) and bottom-right (BR) corners. 

\subsection{Training details}

During training, the coordinates of each corner are converted to gaussian heatmaps of size 512x512, as targets of model's outputs, to allow computing gradients. Each output channel is assigned to a specific corner, in order to be able to recover documents orientation after prediction. We perform data augmentation by randomly applying the following random operations: crop, resizing, rotation, perspective transformation, global illumination scaling, gaussian noise addition. Finally, we resize all images to 512x512 pixels, 3 channels.

\begin{figure}[ht]
  \centering
   \subfloat[After 1 epoch]{\includegraphics[width=.3\textwidth]{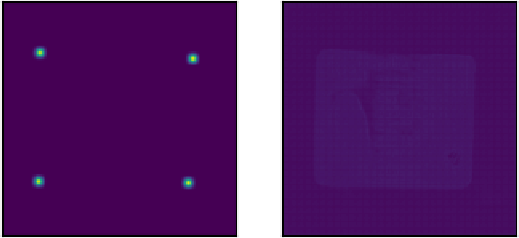}\label{fig:extracted_features_a}}
   \quad
   \subfloat[After 10 epochs]{\includegraphics[width=.3\textwidth]{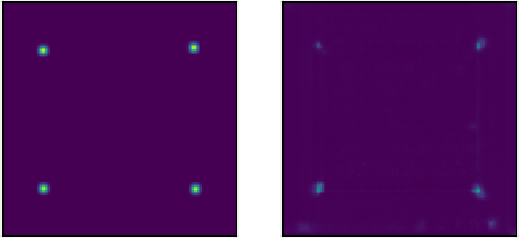}\label{fig:extracted_features_b}}
   \quad
   \subfloat[After 100 epochs]{\includegraphics[width=.3\textwidth]{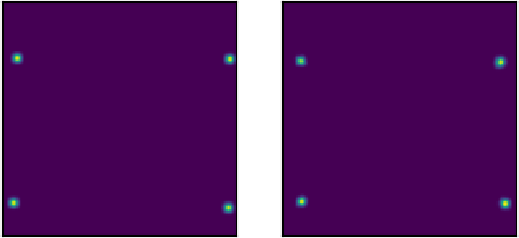}\label{fig:extracted_features_c}}
   \quad \\
   \quad \\
  \caption{Examples of extracted features by model during training (left: labelled points heatmaps, right: predicted points heatmaps). Three different training samples are shown.}
  \label{fig:extracted_features}
\end{figure}

The models are implemented in TensorFlow \cite{abadi2015tensorflow}. Adam Optimizer \cite{kingma2014adam} is used to minimize the MSE loss over the heatmaps, with a learning rate of $2e-4$. All models are trained for 1000 epochs, with an early stopping criterion on validation sub-sets to prevent over-fitting.
The models were trained on 4 GPU cards (two TITAN X cards, and two GTX 1080 Ti cards). Examples of model outputs during training are presented in Figure \ref{fig:extracted_features}.

\section{Experiments}\label{sec:experiments}
The first question we address is how to split the entire model into encoder and decoder parts. We aim to find a good tradeoff between, on the one hand, having an encoder that shares many parameters allowing fast decoder fine-tuning, but on the other hand, having a decoder with enough free parameters to be able to be finetuned and performing well on new document classes.
\subsection{Encoder-Decoder architecture split}
\label{sec:model_split}

We identify four possible encoder-decoder splits, as shown in Figure~\ref{fig:unet_with_split} :
\begin{itemize}
  \item Middle (0): all four upsampler blocks are included in the decoder (like in original U-Net)
 \item 1-, 2-, 3-upsampler(s): one, two or three upsampler blocks are included in the encoder
\end{itemize}

In case of 3-upsamplers split, only one left upsampler block stays in the decoder. Another possible split would be to include all four upsampler blocks in the encoder, with the decoder containing only the transposed convolution layer, but this configuration is excluded from our experiments because it leaves too few parameters in the decoder to learn the task correctly. The different splits are evaluated following Algorithm \ref{lst:training_algorithm}. The encoder weights are frozen during fine-tuning stage. In total, 25 models have been trained, and the following metrics have been computed: 
\begin{itemize}
  \item \textit{IoU} (also referred to as \textit{Jaccard Index} or \textit{Jaccard's coefficient of similarity}): Intersection over Union between ground truth and predicted quadrangles. $\textrm{IoU}(G, P) = \frac{\textrm{A}(G \cap P)}{\textrm{A}(G \cup P)}$. 
  \item \textit{score}: The maximum value of the output channel (heatmap), that we treat as a confidence score for each corner.
\end{itemize}

\begin{algorithm}[ht]
  \centering
  \begin{algorithmic}
    \For{\texttt{\textit{class} in \textit{5 document classes}}}
    \State \texttt{Train full SDL-Net model (M) on all training samples,}
    \State \hspace{.3cm} \texttt{except on those belonging to \textit{class}.}
    \State \texttt{Evaluate M on the holdout test set of \textit{class}.}
    \For{\texttt{\textit{architecture} in 4 encoder-decoder splits}}
    \State \texttt{Initialize a new SDL-Net (M') with the weights of M.}
    \State \texttt{Fine-tune the decoder part of M' on \textit{class}}
    \State \hspace{.3cm} \texttt{while the encoder part of M' is frozen.}
    \State \texttt{Evaluate M' on the test set of \textit{class}.}
    \EndFor
    \EndFor
  \end{algorithmic}
  \caption{Evaluation of split candidates}
  \label{lst:training_algorithm}
\end{algorithm}

Each model produces four probability heatmaps (one for each corner), where each pixel represents the probability of the corresponding corner being located at the pixel. The coordinates of the corner points are finally estimated by extracting the activation peak of each heatmap.

\begin{figure}[ht]
  \centering
  \subfloat[IoU]{\includegraphics[width=0.4\textwidth]{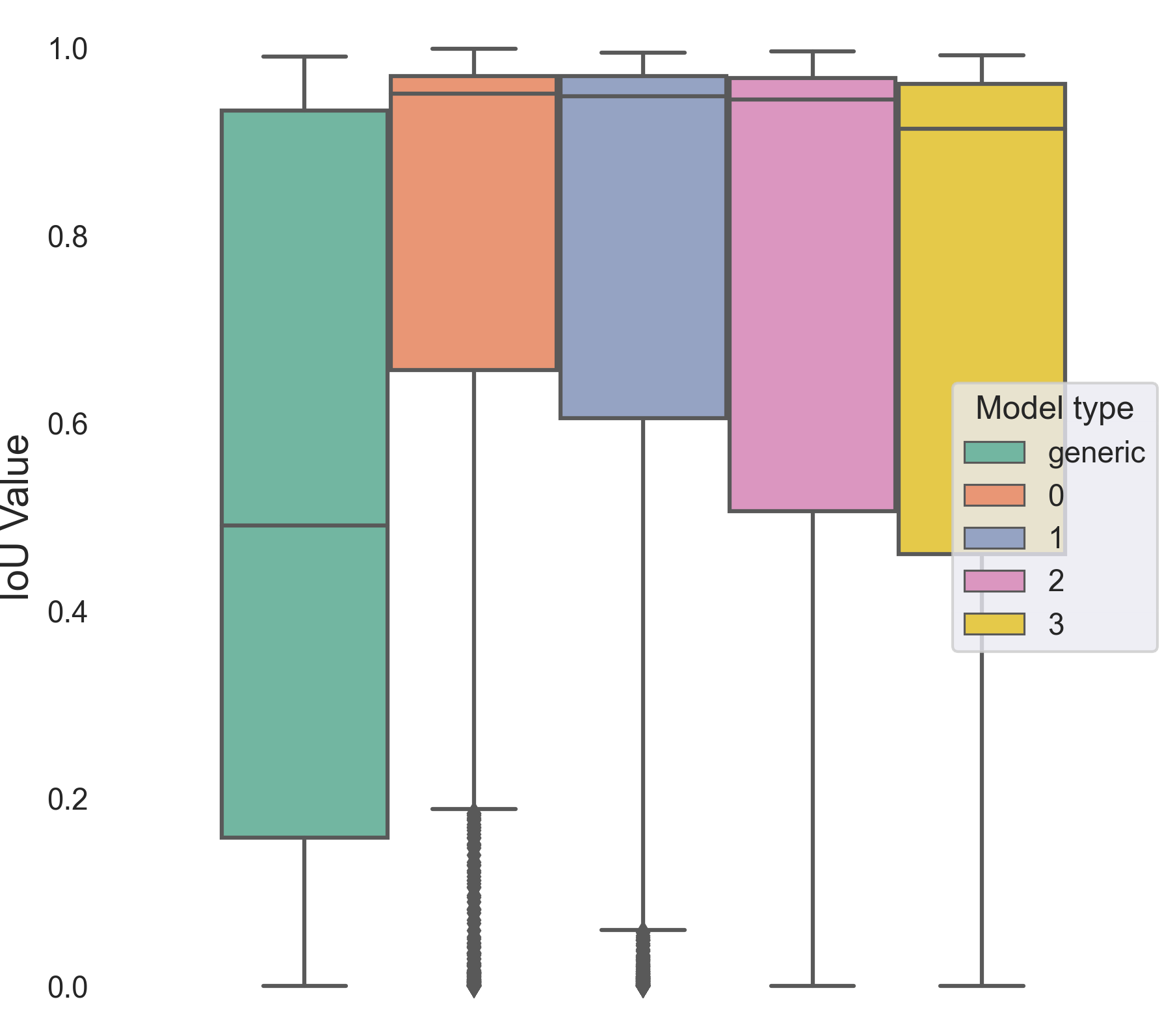}\label{fig:segmentation_metric}}
  \quad
  \subfloat[Confidence score]{\includegraphics[width=0.4\textwidth]{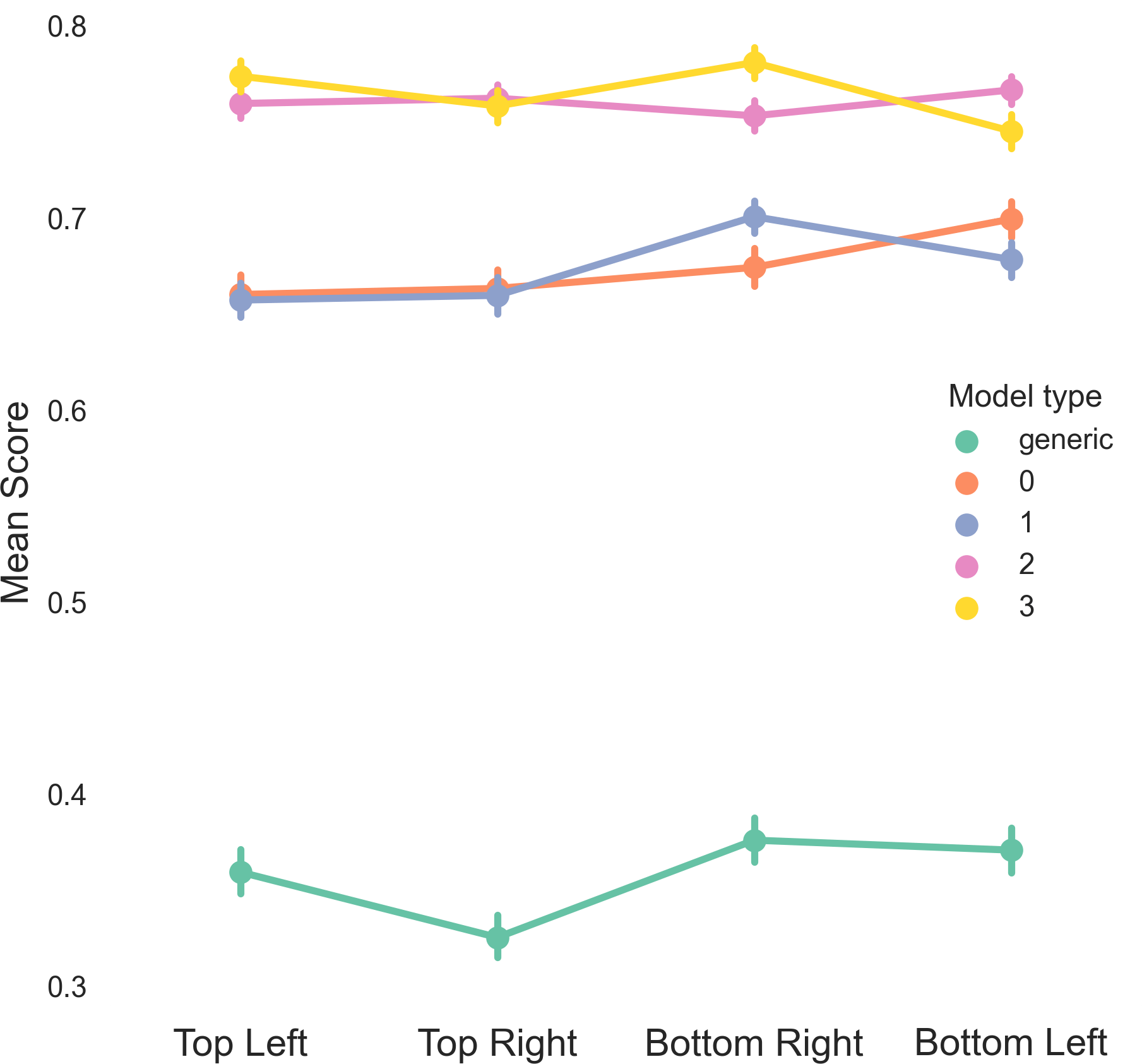}\label{fig:score_metric}}
  \quad \\
  \quad \\
  \caption{Evaluation of generic and fine-tuned models for each encoder-decoder split trained on the four document classes and tested on the hold-out document class.}
  \label{fig:part_1_metrics}
\end{figure}

The plots in Figure \ref{fig:part_1_metrics} show the results of models evaluation, using five different splits, on their respective holdout test dataset. \textit{Generic} corresponds to the model trained on all training datasets except the hold-out document class. The other split types (0, 1, 2, 3) correspond to the models where the encoder was initialized and frozen using weights of the generic model, and the rest (decoder) was finetuned on the hold-out training dataset. 

\begin{figure}[ht]
  \centering
  \includegraphics[width=0.4\textwidth]{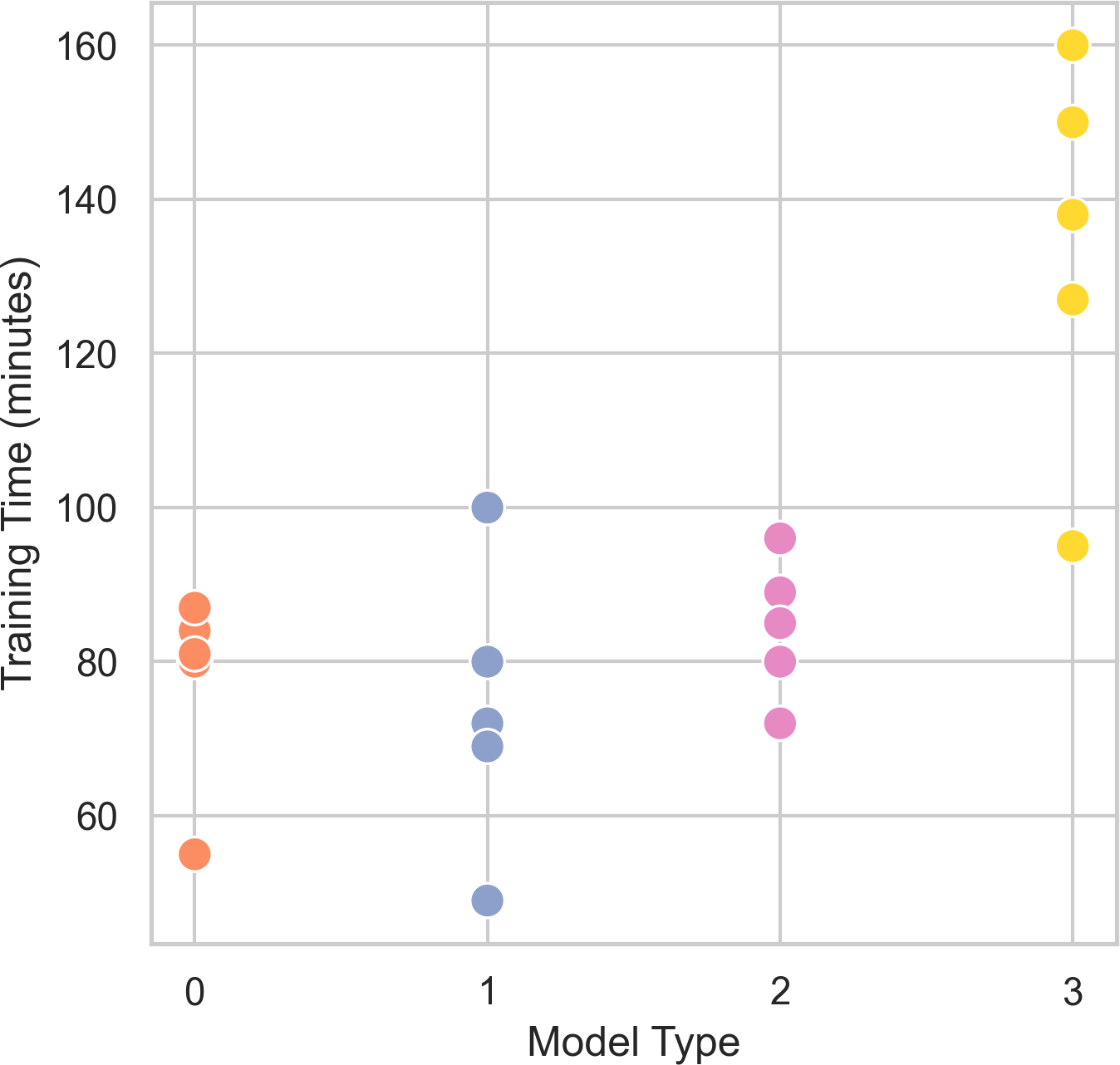}
  \caption{Model training time by architecture encoder-decoder split.}
  \label{fig:training_time}
\end{figure}

The presented metrics show small performance difference between 0 and 1 upsampler splits. To select the best model architecture for our needs, we also evaluate the entire model training time till early stopping in Figure \ref{fig:training_time} for each encoder-decoder split and each document class.

The last model with 3 upsamplers in the encoder, as shown in Figure \ref{fig:training_time}, contains the least parameters in the decoder sub-part, however, it takes more time than others to train. It could be explained by the limited capacity of the model to converge during fine-tuning since the decoder has not enough free parameters to adapt to the new document class. Given all those results, we choose to continue with the split after the first upsampler block (1) since it exhibited good performance vs training time balance.
\subsection{Generalization Power}

We now study the generalization power of the selected architecture. In practice, we study the performance gain obtained by fine-tuning the decoder as a function of the number of document classes used during encoder pre-training phase. We measure the marginal value of fine-tuning and, thus, the generalization power of our encoder: the smaller the marginal value, the higher the generalization power. 

In order to keep the number of experiments reasonable, we arbitrarily select Driving Licence (DL) as the holdout class for this part. We use it only during the fine-tuning and evaluation phases. With the other document classes, we create all possible combinations of one, two, three, or four distinct classes. On each combination, we pre-train the SDL-Net model for 500 epochs, with early stopping. In addition, we carry out two complementary experiments :
\begin{enumerate}
  \item training a model from scratch on the DL class only
  \item training another model on all training datasets
\end{enumerate}

\begin{figure}[!h]
  \centering
   \subfloat[IoU]{\includegraphics[width=.4\textwidth]{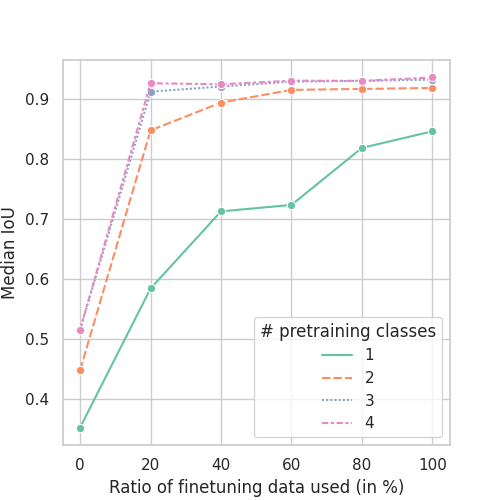}\label{fig:iou_test_subsets}}
   \quad
   \subfloat[Score]{\includegraphics[width=.4\textwidth]{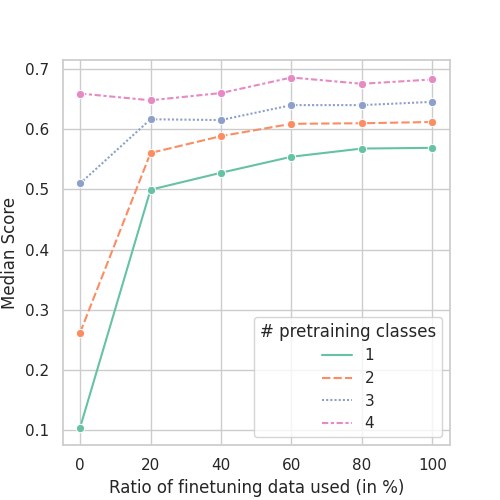}\label{fig:score_test_subsets}}
   \quad \\
   \quad \\
  \caption{Evaluation results on the holdout class test set. The horizontal axis corresponds to the proportion (\%) of the holdout class train set used for fine-tuning.}
  \label{fig:test_subsets_metrics}
\end{figure}

We obtain 17 pre-trained models in total. We fine-tune a specific decoder to localize DL document corners, for each pre-trained encoder. In addition, each decoder was fine-tuned using increasing proportion (20, 40, 60, 80, and 100\%) of the holdout DL train set. We evaluate both pre-trained and fine-tuned models on the holdout DL test set. The evaluation results of those models are presented in Figure \ref{fig:test_subsets_metrics}.

In the following, we report metrics (as $\mu$=mean, $\sigma$=standard deviation, $m$= median), on the holdout class test set for each pre-training combination-size and each ratio of DL dataset used for fine-tuning, e.g. metrics for combinations of one dataset with 20\% of holdout DL dataset correspond to the test metrics of models pre-trained separately on Identity Card, Passport, Vehicle Registration Certificate, and then fine-tuned on 20\% of Driving licence card datasets.

Let's consider training the end-to-end model only on DL images as the baseline. With this approach, we obtain a model that struggles to generalize even on images of the same document class (IoU: $\mu=0.58$, $\sigma=0.38$, $m=0.59$) , as opposed to the model pre-trained on all four other document classes and fine-tuned on DL dataset (IoU: $\mu=0.73$, $\sigma=0.31$, $m=0.94$). We conclude that with a limited number of images in the training dataset, an end-to-end approach trained from scratch does not allow for obtaining an accurate localization model.

The evaluation metrics in Figure \ref{fig:test_subsets_metrics} clearly show that a new unknown document is better localized when more different document classes are used during pre-training stage. It also reveals that models pre-trained on multiple various document classes achieve better results, even with fewer labelled documents used during fine-tuning. It allows us to conclude that increasing the number of document classes used during pre-training allows reducing significantly the need for labelled images during fine-tuning, and makes the model converge faster.
\section{Conclusion}\label{sec:conclusion}

This paper presented an encoder-decoder architecture that allows to pre-train a model for document localization. We performed exhaustive experiments to explore the different possible encoder-decoder splits, and chose the one that was most suited to our use case, with a compromise on the model performance and the time necessary for fine-tuning. 

We evaluated the proposed architecture on multiple combinations of different document classes, which allowed us to explore the generalization capabilities of our model on new documents. Extensive experimental results have demonstrated that we could take advantage of pre-training on diverse datasets with various document classes, by needing only a small number of labeled data when fine-tuning.

Finally, we showed that in the context of industrial constraints, namely the availability of data on rare documents, we can increase the performance of a localization task by sharing a common backbone and leveraging weights trained on corner detection for documents on which data is numerous. 
\section*{Acknowledgments}

We extend our heartfelt gratitude to Alexis Berges and Frédéric Guay for their proposal, organization, and  management of this project, all of which laid the foundation for this research.

Our appreciation also goes to QuickSign for granting us access to computational resources, which proved instrumental for our experiments. We are especially thankful to Florian Praden and Nicolas Girardi for their unwavering technical support and assistance.
\bibliographystyle{splncs04}
\bibliography{references}

\end{document}